# Solving the Steiner Tree Problem in graphs with Variable Neighborhood Descent

Matthieu De Laere[1], San Tu Pham[2], Patrick De Causmaecker[3]

[1] KU Leuven KULAK
Dept. of Computer Sciences
matthieu.delaere@student.kuleuven.be

[2] KU Leuven KULAK
Dept. of Computer Sciences
san.pham@kuleuven.be

[3] KU Leuven KULAK
Dept. of Computer Sciences
patrick.decausmaecker@kuleuven.be

**Abstract**

The Steiner Tree Problem (STP) in graphs is an important problem with various applications in many areas such as design of integrated circuits, evolution theory, networking, etc. In this paper, we propose an algorithm to solve the STP. The algorithm includes a reducer and a solver using Variable Neighborhood Descent (VND), interacting with each other during the search. New constructive heuristics and a vertex score system for intensification purpose are proposed. The algorithm is tested on a set of benchmarks which shows encouraging results.

## 1  Introduction

The Steiner Tree Problem (STP) is an important problem in combinatorial optimization which has numerous applications, ranging from the design of (very large) integrated circuits to computer networking, evolution theory in biology and more [8]. There are plenty variants of the STP which can be found in [7]. The common part between different variants is the requirement to connect a set of objects with the shortest interconnect possible. In this paper, we investigate the general STP in graphs.

As the STP is $\mathcal{NP}$-hard [10], most of the work in the literature focuses on non-exact approaches. Borradaile et. al. [2] proposed polynomial time approximation schemes for the STP in Planar Graphs. A Hybrid Local search approach was contributed by Aragao and Ribeiro [1]. Dowsland worked on Simulated Annealing [4]. An advanced algorithm which includes complicated pre-processing and the combination of Linear Programming formulations and relaxations was proposed by Polzin [14] in his PhD thesis and is the state-of-the-art for the STP.

In this paper, we propose a new algorithm for the STP. The proposed algorithm includes two components: (1) A reducer including a number of tests which reduces the problem size by removing nodes and edges whose removal does not affect the optimal solution. (2) A solver which uses a Variable Neighborhood Descent (VND) algorithm. Two constructive heuristics are proposed to find the initial solution for the VND. For the VND algorithm, two neighborhood structures are defined. During the search, the two components (reducer and solver) communicate with each other to improve their performance. Results obtained by our algorithm are promising, especially on computer chip design related graphs and hypercubes.

The contributions of this paper are threefold. Firstly, the communication between the reducer and the solver is explored. This idea has been mentioned before [14] but has not been receiving enough attention. The communication between two components helps to accelerate both of them. Secondly, two new constructive heuristics are proposed, based on the SDISTG heuristic [16]. Lastly, the vertex score system is proposed for intensification purpose.

The rest of the paper is organized as follows. The problem description is presented is section 2. Section 3 gives an overview about the algorithm framework. The two components of the algorithm are presented in section 4 and section 5. Experimental results are shown in section 6 and section 7 concludes the paper.

## 2 Problem description

Given an undirected graph $G(V, E)$ with a set of nodes $V_G$ and a set of edges $E_G$ (subscript is omitted when the graph is known from the context), the problem consists of finding a subgraph $S$ of $G$ with minimal cost which contains at least all nodes of a set $T \subseteq V$ (called terminals). It is allowed to add other nodes than those from $T$ to $S$ which are called Steiner nodes. $S$ should be a tree, which means that from every node $s$ and $t \in V_S$ there should exist exactly one path between $s$ and $t$.

Throughout this paper, certain definitions and notations are used. The cost of an edge between nodes $i$ and $j$ is referred to as $c_{ij}$. The degree of a node $i$ is $\delta(i)$ and is equal to the number of edges connected to $i$. The Minimum Spanning Tree of a graph is denoted by MST. The distance between nodes $i$ and $j$ is denoted with $d(i, j)$. A distance graph $D_G$ of a graph $G$ is a complete graph containing all nodes of $G$ and all edges have a cost equal to $d_{ij}$ in $G$, for all $i, j \in V$. The terminal distance graph is a subgraph of the distance graph containing only the terminals. With Terminal Minimum Spanning Tree (TMST), we denote the MST of the terminal distance graph. The bottleneck of a path is the maximum cost of all edges in the path.

## 3 Algorithm overview

The algorithm includes two components communicating and interacting with each other during the search. The first component is the reducer which includes a number of reduction tests in order to simplify the graph in a way that does not affect the optimal solution. The second component is the solver which actually solves the problem using a Variable Neighborhood Descent (VND) algorithm. The detailed description of those two components is presented in section 4 and 5 respectively. During the search, the reducer and the solver interchange information to improve the performance of both components. Evidently, removing nodes and edges simplifies the task of the solver. In the other direction, some reduction tests can remove more nodes or edges if a tighter upper bound (given by the solver) is known.

The detailed framework is as follows. At first, using a simple and fast test, the reducer simplifies the graph from which our constructive heuristic constructs the initial solution. This initial solution is then fed to the VND loop. While the initial solution is constructed and the VND runs, the other reduction tests in the reducer run in parallel. Every time a tighter upper bound is found by the VND, it is fed to the bound-based tests. With this new bound, those tests might be able to reduce the graph furthermore which in turn reduces the search space and hence speeds up the VND.

The overview and interaction is illustrated in Figure 1.

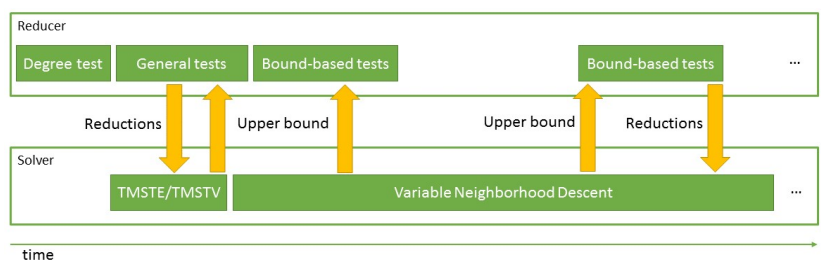

Figure 1: Overview of the algorithm and the interactions.

## 4 Reducer

An instance of the Steiner Tree Problem can often be reduced using a range of different reduction tests, which are briefly described here. Reductions can be interesting since it can reduce the problem size considerably. However, some reduction tests are quite time consuming so there should a selection of tests to be applied. It is also important to note that a reduction does not affect the optimal solution. For details and proofs of these reduction tests, we refer the reader to corresponding publications.

### 4.1 General reduction tests

A first series of tests are those that are universally valid and do not depend on an upper or lower bound for the problem. These tests can be run and reduce the problem as soon as the graph is known. The tests that are run in the algorithm described in this paper will be shortly introduced to the reader.

*Degree test* described in [11], is the simplest reduction test. This test eliminates nodes from the problem graph by using their degree. For example, a node $i \notin T$ with $\delta(i) = 1$ will never be included in an optimal solution. Imagine a solution containing a Steiner node of degree one, then this solution can be improved upon by removing this node and its incident edge, which reduces the cost. This means that a Steiner node of degree one can never be included in an optimal solution.

*Special Distance Test* introduced by Duin and Volgenant in [5] is more sophisticated. To understand the principles of this test, it is convenient to introduce following terminology: a special path is a path in $D_G$ of which all intermediary vertices (if any) are terminals. The special distance between two vertices $i$ and $j$ is the minimum bottleneck over all special paths from $i$ to $j$. This test uses the special distance $s_{ij}$ between two nodes $i$ and $j$ and states that if an edge exists between $i$ and $j$, and $c_{ij} > s_{ij}$, the edge can be removed. The test uses a labeling procedure to find the edges which can be removed. In its original form, the test is very time consuming. Polzin introduced a simplified but faster version in [14]. In the rest of this paper the Special Distance test refers to this simplified version by Polzin, unless explicitly stated otherwise. For more information about this test, see [14].

*Triangle Test* also described by Polzin in [14]. This test works fast and can sometimes be very effective. The test determines $\hat{S}$: the cost of the most expensive edge in the TMST, and removes all edges in $G$ with $c_{ij} > \hat{S}$.

### 4.2 Bound-based reduction

Another series of reduction tests can be run if an upper bound is known for solutions to the instance of the STP. These tests are used to further reduce the problem after finding a (non-optimal) solution to the given problem.

*Reachability Test*, introduced by Duin in [6], states that given an upper bound $B$ for a solution to the problem, a vertex $v$ can be removed if $\max\{d(v, t)\} > B$, for all $t \in T$. This test can be easily run by first computing the distance from every $t$ to all other nodes, which takes $O(|T||E| + |T||V|\log|V|)$. This order of complexity is obtained by running the Dijkstra algorithm $|T|$ times. With these distances available, the only thing needed is to find the maximum distance for every $v$, which takes $O(|V||T|)$.

Another test is the *Voronoi Region* (VR) test, introduced by Polzin in [14]. One can split the graph in different Voronoi regions. Each terminal $z$ defines its own Voronoi region containing all nodes for which this terminal is the closest terminal to the node: $N(z) = \{v \mid d(v,z) < d(v,t),\ t \in T \setminus z,\ v \in V\}$. The terminal that defines the Voronoi region is called the base. The radius of a Voronoi region is the distance from its base to the closest node that is not in the Voronoi region: $radius(z) = \min\{d(z,v) \mid v \notin N(z)\}$. We denote the $p$-th closest terminal to a node $v_i$ by $z_{i,p}$. Assume a graph contains $r$ Voronoi regions, then the radiuses can be ordered as $radius(z_1),\ldots,radius(z_r)$ in ascending order. The VR test states that if $S$ is an optimal Steiner tree and $v_i$ is a Steiner node in $S$, then $d(v_i, z_{i,1}) + d(v_i, z_{i,2}) + \sum_{t=1}^{r-2} radius(z_t)$ is a lower bound for the weight of $S$. With a certain upper bound $B$ available, it is easy to see that nodes $v_i$ for which $d(v_i, z_{i,1}) + d(v_i, z_{i,2}) + \sum_{t=1}^{r-2} radius(z_t) > B$, $v_i$ can impossibly be in an optimal solution and thus can be removed.

## 5 Variable Neighborhood Descent solver

To solve the STP, the Variable Neighborhood Descent (VND) [13] is used. This metaheuristic is widely used in many combinatorial optimization problems [3] [15] [17].

The (VND) is a local search metaheuristic for solving combinatorial optimization problems which systematically changes neighborhoods throughout the local search procedure. The idea behind VND is that a local minimum for one neighborhood is not necessarily a local minimum for another neighborhood. The goal of VND is to traverse these local minima by switching to a different neighborhood. We define two neighborhood structures, each of which has different variants regarding to the size of the neighborhood. Before the VND procedure can start, an initial solution is needed. In the below subsections, all components of the VND are presented including the initial solution construction (section 5.1), the vertex scores that are utilized in the local search (section 5.2) and two local search procedures corresponding to two neighborhood moves (sections 5.3, 5.4).

---

**Algorithm 1:** Variable Neighborhood Descent for the STP

**Require:** Graph $G(V, E)$, min and max neighborhood size $B_{min}$ and $B_{max}$
  $b \leftarrow B_{min}$
  $S \leftarrow InitialSolution()$
  $S_{best} \leftarrow S$
  **while** $B < B_{max}$ **do**
    $S \leftarrow InsertionLocalSearch(S, B)$
    **if** solution has improved **then**
      $S_{best} \leftarrow S$
      $b \leftarrow B_{min}$
    **else**
      $Rem \leftarrow RemovalLocalSearch(S, B)$
      **if** solution has improved **then**
        $S_{best} \leftarrow S$
        $b \leftarrow B_{min}$
      **else**
        $b \leftarrow b * 2$ // Increase the neighborhood size
      **end if**
    **end if**
  **end while**

---

### 5.1 Constructive algorithms

Any graph that is a tree and contains all nodes from $T$ would suffice as an initial solution for the local search algorithm to start. That means the MST of the graph makes a valid intial solution which is easily obtained using Kruskal's algorithm. However, its quality is generally not good.

In this section, we propose two simple constructive algorithms to construct initial solution for the local search procedures. They are based on the SDISTG heuristic from Voß[16]. SDISTG works by creating the TMST and expanding the edges by the paths they represent. Then in the original graph "edges and Steiner nodes are removed such that no cycles exist and all leaves are terminals". However, it does not state how to reach this. We propose two different ways of achieving this last step.

Both of our two constructive algorithms begin with the construction the "Terminal Shortest distance graph" where set of vertices contains the terminals of the original graph and edge weights represent the shortest distance also in the original graph. Then we obtain the the MST of this graph which is called TMST. The TMST is then expanded with the paths from the original graph. In the pruning step, which is where the two algorithms differ, the original graph is pruned based on the TMST. In the first algorithm, all the edges that are not in the TMST are removed from the original graph. This is called Terminal

Minimum Spanning Tree with Edge pruning (TMSTE). Another possibility is to remove all vertices that are not in the TMST. This is the TMSTV algorithm (Terminal Minimum Spanning Tree with Vertex pruning). From each of those graphs, we take the MST to remove possible cycles, which leads to two solutions. The best of these two is the starting point for the local search algorithm. The algorithm is illustrated in Figure 2.

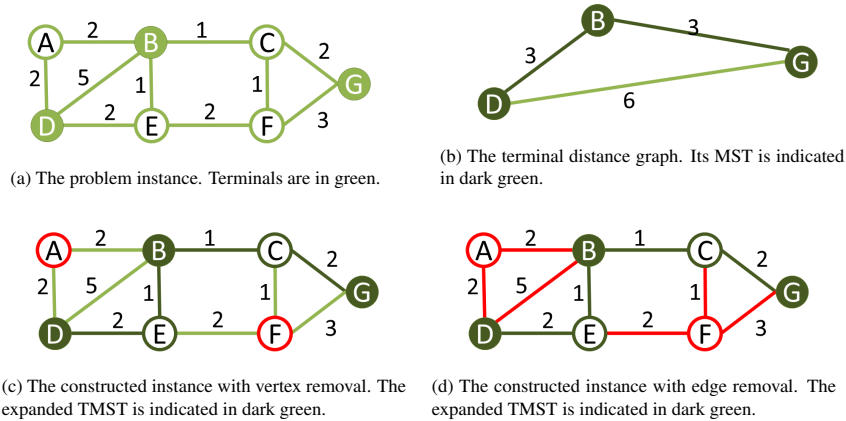

(a) The problem instance. Terminals are in green.

(b) The terminal distance graph. Its MST is indicated in dark green.

(c) The constructed instance with vertex removal. The expanded TMST is indicated in dark green.

(d) The constructed instance with edge removal. The expanded TMST is indicated in dark green.

Figure 2: The construction of an initial solution.

### 5.2 Vertex scores

For intensification purpose, the concept of score is introduced. Each node gets a score which can intuitively be seen as how much a certain node is preferred to be in the solution of the STP. At the initialization, terminals get the highest score and other nodes are neither preferred nor ignored. The scores for terminals are immutable. Throughout the search process the score of one node is increased if its presence improves the current solution and decreased otherwise.

This score system is integrated into the local search procedure as follows. During the local search process, when a new solution is found, the scores of newly inserted nodes will be decreased if the new solution is worse than the previous one. Otherwise, scores of all nodes in the new solution are increased to reward their staying in a good solution. The scores for each node are kept tracked and the average value within 10 most recent iterations is considered in the local search procedure.

To avoid over-scored situations, random restarts are applied to the score system. During the VND procedure, when the maximum neighborhood size is reached, the scores in the graph get a random re-initialization.

### 5.3 Insertion local search

The Steiner node insertion local search works by taking a random starting node $i$ from the current solution with $\delta(i) > 2$ and navigating a path until another node from the current solution $j$ with $\delta(j) > 2$ is met. This path then gets inserted in the current solution, while maintaining the tree property. Lastly, the solution is pruned by removing Steiner nodes of degree one. If this leads to a better solution, it is accepted; if not, this new (worse) solution is used to navigate to the next neighbor.

In these neighborhoods the parameter $B$ influences the number of starting nodes, and as a consequence the number of paths added to the graph. The start nodes are taken based on the scores information, i.e. the first $B \ln^2 V$ best-scored nodes are chosen.

---

**Algorithm 2:** Insertion local search
**Require:** Graph $G(V, E)$, set of terminals $T \subseteq V$, current solution $S(V, E)$ and neighborhood size parameter $B$.
  **while** Number of random score restarts is not reached **do**
    $StartNodes \leftarrow$ Take the $B \ln^2 V$ first nodes $i$ with $\delta(i) > 2$ ordered by descending average score
    $S' \leftarrow S$
    **for all** nodes in $StartNodes$ **do**
      $P \leftarrow$ Random path to another node $s$ in $S$ with $\delta(s) > 2$
      $S' \leftarrow$ Insert edges of $P$ into $S'$ while keeping the three property
      Prune nodes with degree one from $S'$
      **if** Cost of $S'$ < Cost of $S$ **then**
        Increase score of all nodes in $S'$
        **return** $S'$ // Return first improvement
      **else**
        Decrease score of all new inserted nodes in $S'$
      **end if**
    **end for**
    Do random restart of the scores
  **end while**
  **return** $S'$

---

### 5.4 Removal local search

The Steiner node removal local search is more time consuming than the insertion local search. It removes Steiner nodes from the current solution and then reconnects the disconnected components. This can be done by randomly taking some nodes from each component and running the Dijkstra algorithm to calculate the shortest distance to every other node in the graph. With this information available, a component connecting graph is created in which each component is represented by one node. The edges between those nodes have the cost of the shortest path found in the previous step. After taking the MST of this graph, the solution can be reconnected by expanding these paths and adding them to the disconnected solution.

The removal local search is described in Algorithm 3. The neighborhood size parameter $B$ is the number of Steiner nodes removed at once. Removing multiple Steiner nodes leads to more disconnected components, which in turn leads to more ways to reconnect.

## 6 Experimental results

To test the algorithm, the SteinLib benchmarks [12] were used. The SteinLib library is divided into different test sets. Each test set in SteinLib is labeled with one of these difficulty ratings:

- **Dead:** These sets of instances have been solved to optimality in various publications. In our results, this category is labeled **D**.

- **Solved:** These sets of instances have been solved to optimality in less than two independent publications. This category is labeled **S**.

- **Unsolved:** These sets of instances have not been solved to optimality in any publication to date. This category is labeled **U**.

Our algorithm was tested on some of the sets. The first is the `E` test set, which contains random generated sparse graphs with edge weights between 1 and 10. The second set is `TAQ`, containing grid graphs with rectangular holes. These graphs are typically used in VLSI applications (Very-Large-Scale

---

**Algorithm 3:** Removal local search
**Require:** Graph $G(V, E)$, set of terminals $T \subseteq V$, current solution $S(V, E)$ and neighborhood size parameter $B$.
　**while** Number of random score restarts is not reached **do**
　　$PossibleVictims \leftarrow$ Take $3B$ first Steiner nodes from $S$ ordered by descending average score
　　$Combinations \leftarrow$ Take a number of combinations of size $B$ from $PossibleVictiims$
　　**for all** combinations in $Combinations$ **do**
　　　$S' \leftarrow S$ without all nodes $i$ in combination from $S$ (so $S'$ is a disconnected graph)
　　　$S' \leftarrow Reconnect(S')$
　　　**if** Cost of $S' <$ Cost of $S$ **then**
　　　　Increase score of all nodes in $S'$
　　　　**return** $S'$ // Return first improvement
　　　**else**
　　　　Decrease score of all new inserted nodes in $S'$
　　　**end if**
　　**end for**
　　Do random restart of the scores
　**end while**

---

Integration, the design of very large integrated circuits) [9]. The last set is the `PUC` test set containing hypercubes. For more information about the SteinLib benchmarks, see [12].

The algorithm was implemented in C# with .NET 4.5. The experiments were run on a machine with an Intel Core i7 3632QM CPU and 12GB RAM, running Windows 10. The parameters settings for the problem are as follows: the minimum and maximum neighborhood sizes are respectively $B_{min} = 1$ and $B_{max} = 256$, the maximum number of random score restarts is 5.

The experiments were run 8 times on each instance. The results are shown in Table 1. The columns show (from left to right): the problem instance, number of vertices, number of edges, number of terminals, the best, average and worst solution found by the algorithm, the average running time to find the best solution, the average gap to the optimal solution, the standard deviation on the solutions found and the best known solution. If this best known solution is optimal, it is printed in bold. In the remaining columns the results achieved by Polzin are mentioned, except for the `TAQ` set as those were not tested on.

The results per instance, grouped per test set, can be found in Table 1 and 2. In these tables, we report the performance of our algorithm and compare our solutions with the best known and the state-of-the-art results obtained by Polzin [14]. An overview with time needed and the average gap per test set can be found in Table 2. The table shows the number of instances for which an optimal solution was found, the average time needed to solve the entire set and the average gap to the optimal for the entire set.

From these results, it is easy to see that the algorithm performs well on the `TAQ` test set, with an average gap of only $1.15\%$. It is interesting to note that on this set, the results seem to be quite consistent, which is affirmed by the low standard deviation. For the `E` test set, results are slightly worse, with an average gap of $1.9\%$. This is interesting as this set is considered easier than the `TAQ` set.

Results on the `PUC` test set are good, considering the time it takes to achieve them. For instance, when comparing the `HC7U` instance, the solution provided by Polzin is optimal, but to achieve it he needs almost 5 hours, while our VND algorithm only has a $0.82\%$ gap in just over one minute even though our algorithm is much simpler than his.

## 7 Conclusion and future work

In this paper we proposed a solver for the STP which includes two components - reducer and solver - communicating with each other during the search. The solver uses the Variable Neighborhood Descent as the solving method. We also contributed two new constructive heuristics for the problem. The vertex score system was proposed to intensify the search. The experimental results are promising, especially on hypercube and VLSI-derived instances. In general, our results are not comparable to the state-of-the-art regarding to solution gaps. However, taking into account the simplicity of our algorithm with the use of very simple reduction tests, those results are promising and there is still a lot of room for improvement.

| Instance | $|V|$ | $|E|$ | $|T|$ | Our VND algorithm | | | | | | Best known | Polzin's algorithm [14] | |
|---|---|---|---|---|---|---|---|---|---|---|---|---|
| | | | | Best | Average | Worst | Time (s) | Gap (%) | Stdev | | Best | Time (s) |
| **E (D)** | | | | | | | | | | | | |
| E01 | 2500 | 3125 | 5 | **111** | 111 | 111 | 10.4 | 0.00 | 0 | **111** | 111 | 0.5 |
| E02 | 2500 | 3125 | 10 | 227 | 227 | 227 | 56.5 | 6.07 | 0 | **214** | 214 | 0.3 |
| E03 | 2500 | 3125 | 417 | 4035 | 4038.3 | 4043 | 2556.6 | 0.63 | 2.31 | **4013** | 4013 | 0.1 |
| E04 | 2500 | 3125 | 625 | 5115 | 5122.2 | 5137 | 3819.9 | 0.42 | 2.31 | **5101** | 5101 | 0.1 |
| E09 | 2500 | 5000 | 625 | 3634 | 3645 | 3662 | 6103.6 | 1.14 | 7.77 | **3604** | 3604 | 0.1 |
| E10 | 2500 | 5000 | 1250 | 5625 | 5630 | 5634 | 6908.8 | 0.54 | 2.53 | **5600** | 5600 | 0.1 |
| E11 | 2500 | 12500 | 5 | **34** | 35 | 37 | 101.8 | 2.94 | 1.41 | **34** | 34 | 0.8 |
| E12 | 2500 | 12500 | 10 | 68 | 68 | 68 | 184.2 | 1.49 | 0 | **67** | 67 | 0.6 |
| E13 | 2500 | 12500 | 417 | 1307 | 1313.3 | 1319 | 2973.7 | 2.60 | 4.35 | **1280** | 1280 | 1.3 |
| E14 | 2500 | 12500 | 625 | 1758 | 1760.1 | 1762 | 3305.3 | 1.62 | 1.6 | **1732** | 1732 | 0.2 |
| E15 | 2500 | 12500 | 1250 | 2799 | 2800.4 | 2805 | 8341 | 0.59 | 2.17 | **2784** | 2784 | 0.2 |
| E16 | 2500 | 62500 | 5 | **15** | 15 | 15 | 0.8 | 0.00 | 0 | **15** | 15 | 0.8 |
| E17 | 2500 | 62500 | 10 | **25** | 25.1 | 26 | 221.2 | 0.44 | 0.31 | **25** | 25 | 0.5 |
| E18 | 2500 | 62500 | 417 | 597 | 598.7 | 602 | 1391.1 | 6.15 | 1.49 | **564** | 564 | 21.6 |
| E19 | 2500 | 62500 | 625 | 782 | 783.6 | 787 | 1446.4 | 3.37 | 1.5 | **758** | 758 | 4.1 |
| E20 | 2500 | 62500 | 1250 | 1349 | 1349.9 | 1353 | 4336.8 | 0.59 | 1.2 | **1342** | 1342 | 0.2 |
| **TAQ (D)** | | | | | | | | | | | | |
| TAQ0014 | 6466 | 11046 | 128 | 5402 | 5446.9 | 5476 | 4438.8 | 2.27 | 23.77 | **5326** | — | — |
| TAQ0023 | 572 | 963 | 11 | **623** | 623 | 623 | 13.3 | 0.32 | 0 | **621** | — | — |
| TAQ0365 | 4186 | 7074 | 22 | 1919 | 1931.6 | 1944 | 653 | 0.92 | 7.27 | **1914** | — | — |
| TAQ0377 | 6836 | 11715 | 136 | 6544 | 6611.7 | 6655 | 5750.3 | 3.42 | 40.66 | **6393** | — | — |
| TAQ0431 | 1128 | 1905 | 13 | **897** | 903 | 924 | 59.6 | 0.67 | 11.22 | **897** | — | — |
| TAQ0631 | 609 | 932 | 10 | **581** | 584.4 | 595 | 11.4 | 0.59 | 5.52 | **581** | — | — |
| TAQ0739 | 837 | 1438 | 16 | **848** | 848.7 | 851 | 18.5 | 0.08 | 1.25 | **848** | — | — |
| TAQ0741 | 712 | 1217 | 16 | 852 | 857.3 | 862 | 20.2 | 1.22 | 3.74 | **847** | — | — |
| TAQ0751 | 1051 | 1791 | 16 | **939** | 940.8 | 952 | 23.3 | 0.19 | 4.08 | **939** | — | — |
| TAQ0903 | 6163 | 10490 | 130 | 5166 | 5191.6 | 5216 | 3840.8 | 1.82 | 15.73 | **5099** | — | — |
| **PUC (U)** | | | | | | | | | | | | |
| HC6P | 64 | 192 | 32 | 4026 | 4093.9 | 4133 | 1.3 | 2.27 | 35.45 | **4003** | 4003 | 27.7 |
| HC6U | 64 | 192 | 32 | **39** | 40 | 42 | 1.6 | 2.56 | 1.05 | **39** | 39 | 13.5 |
| HC7P | 128 | 448 | 64 | 7924 | 8020.6 | 8106 | 20.5 | 1.46 | 69.10 | **7905** | 7905 | 14362.7 |
| HC7U | 128 | 448 | 64 | **77** | 78.7 | 80 | 12.7 | 2.16 | 0.82 | **77** | 77 | 17253.5 |
| HC8P | 256 | 1024 | 128 | 15680 | 15888.8 | 16104 | 146.2 | 3.70 | 114.27 | **15322** | 15327 | >18000 |
| HC8U | 256 | 1024 | 128 | 154 | 156.9 | 160 | 76.2 | 6.01 | 2.33 | **148** | 148 | >18000 |
| HC9P | 512 | 2304 | 256 | 31458 | 31548.8 | 31757 | 1198.1 | 4.32 | 94.65 | **30242** | 30310 | >18000 |
| HC9U | 512 | 2304 | 256 | 308 | 310.7 | 317 | 512 | 6.39 | 2.75 | **292** | 292 | >18000 |
| HC10P | 1024 | 5120 | 512 | 62232 | 63853.8 | 63322 | 7210.9 | 5.11 | 337 | **59797** | 60679 | >18000 |
| HC10U | 1024 | 5120 | 512 | 613 | 633.6 | 652 | 2506.4 | 10.18 | 12.47 | **575** | 581 | >18000 |

Table 1: Test results for the VND algorithm on different instances, grouped by test set and compared to results from Polzin.

Regarding to future work, we observe that our solver works well on some types of graphs (like the hypercubes or VLSI-derived ones), while unable to make any move on others, such as the random complete graphs with Euclidean weights. More research is needed to determine on which type of graph our approach is preferable. We are also interested in further exploring the communication between the reducer and the solver. Using more sophisticated reducing algorithms will lead to more pruning during the search and the win-win relationship between the two components will be accelerated.

| Test set | Instances | Optima | Time (s) | Gap (%) |
|---|---|---|---|---|
| E | 19 | 6 | 51134.7 | 1.9 |
| TAQ | 10 | 4 | 14829.1 | 1.15 |
| PUC | 10 | 2 | 11685.8 | 4.42 |

Table 2: Test results for the VND algorithm summarized per test set, only containing the instances which can not be solved by preprocessing, compared to results obtained by Aragao and Ribeiro.

Barcelona, July 4-7, 2017

## References

[1] Mp De Aragao and Cc Ribeiro. Hybrid local search for the Steiner problem in graphs. *Extended abstracts of . . .* , (i):429–434, 2001.

[2] Glencora Borradaile, Philip Klein, and Claire Mathieu. An O ( n log n ) Approximation Scheme for Steiner Tree in Planar Graphs. V, 1999.

[3] Shaowei Cai, Kaile Su, and Abdul Sattar. Local search with edge weighting and configuration checking heuristics for minimum vertex cover. *Artificial Intelligence*, 175(9):1672–1696, 2011.

[4] Kathryn A Dowsland. Hill-climbing, simulated annealing and the Steiner problem in graphs. *Engineering Optimization*, 17(1-2):91–107, 1991.

[5] C. W. Duin and A. Volgenant. An edge elimination test for the Steiner problem in graphs. 8:79–83, 1989.

[6] Cees Duin. Preprocessing the steiner problem in graphs. In *Advances in Steiner Trees*, pages 175–233. Springer, 2000.

[7] M Hauptmann and M Karpinski. A Compendium on Steiner Tree Problems. pages 1–44, 2013.

[8] Frank K Hwang, Dana S Richards, and Pawel Winter. *The Steiner tree problem*, volume 53. Elsevier, 1992.

[9] M. Jünger, A. Martin, G. Reinelt, and R. Weismantel. Quadratic 0/1 optimization and a decomposition approach for the placement of electronic circuits. *Mathematical Programming*, 63:257–279, 1994.

[10] Richard M Karp. *Reducibility among combinatorial problems*. Springer, 1972.

[11] Thorsten Koch, T Koch, and A Martin. Solving Steiner Tree Problems in Graphs to Optimality. 42(December), 1996.

[12] Thorsten Koch, Alexander Martin, and Stefan Voß. *SteinLib: An updated library on Steiner tree problems in graphs*. Springer, 2001.

[13] Nenad Mladenović and Pierre Hansen. Variable neighborhood search. *Computers & Operations Research*, 24(11):1097–1100, 1997.

[14] Tobias Polzin. *Algorithms for the Steiner problem in networks*. PhD thesis, Universitätsbibliothek, 2003.

[15] Bart Selman, Hector J Levesque, David G Mitchell, et al. A new method for solving hard satisfiability problems. In *AAAI*, volume 92, pages 440–446, 1992.

[16] Stefan Voß. Steiner's problem in graphs: heuristic methods. *Discrete Applied Mathematics*, 40(1):45–72, 1992.

[17] Christos Voudouris and Edward Tsang. Guided local search and its application to the traveling salesman problem. *European journal of operational research*, 113(2):469–499, 1999.